# REACT: Two Datasets for Analyzing Both Human Reactions and Evaluative Feedback to Robots Over Time

Kate Candon
Yale University
New Haven, CT, USA
kate.candon@yale.edu

Nicholas C. Georgiou
Yale University
New Haven, CT, USA
nicholas.georgiou@yale.edu

Helen Zhou
Yale University
New Haven, CT, USA
helen.zhou@yale.edu

Sidney Richardson
Yale University
New Haven, CT, USA
sidney.richardson@yale.edu

Qiping Zhang
Yale University
New Haven, CT, USA
qiping.zhang@yale.edu

Brian Scassellati
Yale University
New Haven, CT, USA
brian.scassellati@yale.edu

Marynel Vázquez
Yale University
New Haven, CT, USA
marynel.vazquez@yale.edu

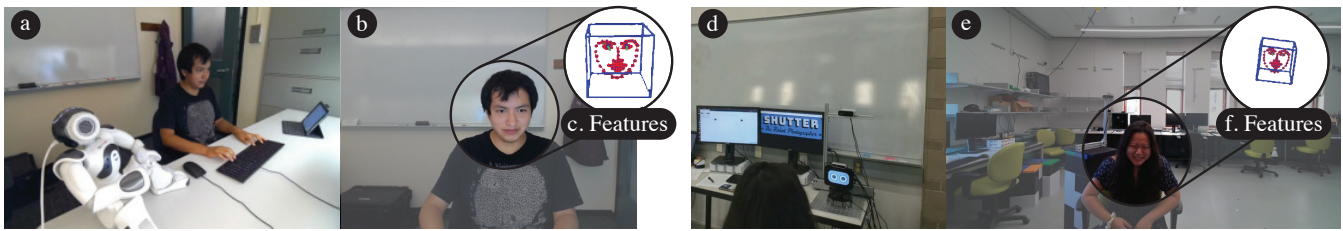

Figure 1: Overview of REACT. In REACT-Nao, people played a collaborative video game with a Nao robot (a). In REACT-Shutter, participants interacted with a Shutter robot during a photography task (d). For both datasets, we captured images of participants throughout the interaction (b,e) and provide facial analyses of the images (c,f).

## ABSTRACT
Recent work in Human-Robot Interaction (HRI) has shown that robots can leverage implicit communicative signals from users to understand how they are being perceived during interactions. For example, these signals can be gaze patterns, facial expressions, or body motions that reflect internal human states. To facilitate future research in this direction, we contribute the REACT database, a collection of two datasets of human-robot interactions that display users' natural reactions to robots during a collaborative game and a photography scenario. Further, we analyze the datasets to show that interaction history is an important factor that can influence human reactions to robots. As a result, we believe that future models for interpreting implicit feedback in HRI should explicitly account for this history. REACT opens up doors to this possibility in the future.

## CCS CONCEPTS
• **Human-centered computing** → *Empirical studies in HCI*; • **Computing methodologies** → *Artificial intelligence.*

## KEYWORDS
human-robot interaction; nonverbal behavior; human feedback

## 1 INTRODUCTION
Robots promise a future where they will help us with many physical and social tasks in human environments. However, as robots enter these environments, such as homes, many tasks will become subjective and driven by personal preferences [7, 25]. Because of this, it becomes infeasible to pre-program all tasks with which we may want robot assistance. Rather, it is essential to make robots better at learning from non-expert human teachers [3].

Human nonverbal reactions are a key and often underutilized source of information for learning from users in Human-Robot Interaction (HRI). Humans naturally convey information through their nonverbal behavior that provides cues about how they perceive social encounters [19, 30]. Indeed, work in affective computing [15, 26] and social signal processing [29] has studied how we can create computational models to interpret human nonverbal reactions. More recently, work in HRI has started to explore this possibility (e.g., [11, 14]). It is generally agreed upon that effective social agents must be able to analyze, comprehend, and respond to nonverbal cues [12]. However, interpreting these cues can be challenging. Different cultures or situations can result in similar nonverbal cues, so these cues may have different meanings depending on the context in which they are generated [5, 9, 17].

In order to facilitate further research on how robots may leverage human nonverbal behavior in HRI, we contribute the Reactions and EvaluAtive feedbaCk over Time (REACT) database. REACT consists of two datasets that contain observations of humans, robots, and task-related data during human-robot interactions (as shown in Figure 1). The first dataset, **REACT-Nao**, consists of data from interactions from a user study [10] in which humans played a video game with a Nao robot while providing explicit feedback so that the Nao could learn to be a better teammate. REACT-Nao includes approximately 864 minutes of data collected across 72 participants. The second dataset, **REACT-Shutter**, consists of observations from interactions with a tabletop social robot during a photography task. REACT-Shutter

*Preprint*

Table 1: Comparison of related available datasets. "Interactive task" indicates whether the human is actively interacting with the robot. "Additional task(s)" indicates if the participant had additional tasks other than just providing feedback to the robot (e.g., playing game in REACT-Nao). "Evaluative feedback" refers to if the dataset includes explicit, evaluative feedback about the robot from the participant throughout the interaction (either live or through annotations). The "Context" columns describe what additional context is provided in the dataset: E = Environment (e.g., location of enemies in REACT-Nao); H = Human (e.g., whether human spaceship moved left or right in REACT-Nao); R = Robot / agent (e.g., actual text of robot utterances in REACT-Shutter).

| Dataset | Nonverbal Features | | | | | Task | | | | Context | | | History |
|---|---|---|---|---|---|---|---|---|---|---|---|---|---|
| | Head pose | Gaze | Facial landmarks | Facial AUs | Raw images | Colocated robot | Interactive task | Additional task(s) | Evaluative feedback | E. | H. | R. | Spans interaction |
| EMPATHIC [13] | ✓ | ✓ | ✓ | ✓ | ✓ | ✗ | ✗ | ✗ | ✗ | ✓ | ✗ | ✓ | ✓ |
| Errors in HRI [28] | ✗ | ✗ | ✗ | ✓ | ✗ | ✓ | ✓ | ✓ | ✗ | ✗ | ✗ | ✗ | ✓ |
| REACT-Nao | ✓ | ✓ | ✓ | ✓ | ✓ | ✓ | ✓ | ✓ | ✓ | ✓ | ✓ | ✓ | ✓ |
| REACT-Shutter | ✓ | ✓ | ✓ | ✓ | ✗ | ✓ | ✓ | ✓ | ✓ | ✓ | ✗ | ✓ | ✓ |

includes approximately 160 minutes of data collected across 40 participants. Part of the latter data was used to investigate different annotation methods of robot performance during interactions [31]. In this work, we augmented this data with additional observations over the whole interaction to provide a more complete dataset to study human implicit signals in HRI. Together, the datasets provide a rich set of observations to analyze how human reactions are related to explicitly provided robot feedback. The datasets and documentation are available at: github.com/yale-img/react.

As a second contribution, we analyze the datasets to evaluate a common assumption in how machine learning models are used to make predictions about users from their nonverbal behavior in HRI. In particular, prior work often focuses on making predictions from short horizons of observations (e.g., [14, 31]). However, our analyses suggest that humans may become less reactive to robots over time. Thus, in the future, it is important for data-driven models to more explicitly account for interaction history in HRI. The data that we contribute in this work opens up possibilities in this respect.

## 2 RELATED WORK

**Existing Datasets.** There is a long history of open datasets with human nonverbal reactions (e.g., see [27] for a survey on human facial expression recognition); however, such datasets are still scarce within HRI. There exist some datasets of human nonverbal reactions to robots [6, 8, 13, 18, 24, 28]. Out of this set, the two publicly available datasets that are closest to REACT involve participants watching robots commit errors during an interactive task [28] and watching agents perform a task sub-optimally [13], as detailed in Table 1. The other datasets [6, 8, 18, 24] provide great value to the field of HRI, but do not facilitate research examining both nonverbal human reactions and explicit evaluative feedback during a task in which both the human and robot play a key role. Our dataset includes both explicit, evaluative feedback and implicit, nonverbal reactions from participants that were actively interacting with a robot during a task. In comparison, the BAD Dataset [8] does not involve humans that are actively interacting with or explicitly evaluating a robot, but rather are reacting to videos that they observe online as bystanders. Similarly, the other datasets [6, 18, 24] do not include explicit feedback during the task. Rather, these datasets support other specific research avenues (e.g., modeling user engagement).

**Reasoning about Human Nonverbal Reactions.** In prior work, models that reason about human nonverbal reactions to robots typically fail to account for a rich interaction history. It is a common approach to reason about nonverbal cues at the individual snapshot level (e.g., [28]), especially when inferring specific emotions or user states (e.g., [20]). Another approach is to examine changes in expressivity over fixed windows (e.g., [22]). While some models incorporate recurrence, they do not explicitly account for how human feedback may change over time (e.g., [31]). Our analyses suggest that as human-robot interactions evolve over time, human nonverbal signals may become more muted, requiring potentially different interpretations based on the interaction history. Going forward, it will be important to investigate algorithms that intelligently reason about feedback that is dependent on other factors, such as a longer interaction history or modeling of internal human states. This type of approach has been explored for reasoning about explicit human feedback, e.g., COACH learns from policy-dependent feedback [23].

## 3 THE REACT-NAO DATASET

The first dataset, REACT-Nao, contains observations throughout a collaborative game between a Nao robot and humans [10].

### 3.1 Data Collection

First, participants consented to take part in the data collection, be video recorded and have their data shared. Participants played six games of Space Invaders with a Nao robot (Figure 1a). They were instructed to provide feedback to the robot via their keyboard during the game so the robot could learn to be a better teammate. In the Space Invaders game, the goal was to destroy all enemies as a team. Each player generally took care of destroying enemies on one side of the game screen. However, the Nao employed different gameplay strategies across games which varied by when the robot's spaceship crossed over to the human's side of the gamescreen to help destroy enemies – we refer to these events as "visits". During games 1 and 2, the robot did not crossover to the human's side to provide assistance. During games 3 and 4, the robot crossed over to the human's side for assistance on three separate "visits". During games 5 and 6, the robot only crossed over for one "visit" at the end of the game, after it had destroyed all of the enemies on its own side.

Participants were not prompted to speak during the interactions, but experimenters noted that some participants did speak at times.

Participants answered survey questions after each pair of games, and a final set of survey questions. The interaction lasted approximately 35 minutes, and the participants were compensated US$10. The protocol was reviewed by our Institutional Review Board (IRB) and refined via pilots. For additional motivations and details of the user study, please refer to the work by Candon et al. [10].

### 3.2 Data Processing

The dataset consists of data collected for 72 participants during the six games of Space Invaders that they each played.

**Facial Features.** To analyze the images captured during the interaction, we used OpenFace 2.0 [4], a open-source toolkit for automatic facial behavior analysis. OpenFace 2.0 [4] uses computer vision algorithms to analyze each image and extract features about head pose, eye gaze, facial landmarks, and facial action units (AUs). Our data is organized in individual CSV files per game and participant. Each CSV file has one row per frame that includes a frame number and the output from running OpenFace on the image from that frame. A detailed description of individual features is included in the dataset documentation.

For our analyses, we first smoothed individual OpenFace features with a Gaussian filter (with a rolling window with a width of 30 data points and a Gaussian function with a standard deviation of 10). We then segmented the frames into "visits" by when the robot's spaceship was on the participant's side of the screen. We examined the mean of values of OpenFace activation values during various "visits" across the games of Space Invaders to see how participants reacted to a change in robot gameplay behavior. All post-processing scripts are included in github.com/yale-img/react.

**Other features.** Our dataset includes additional information that provides context about the interaction. For each game, we provide a json file that contains game state information, robot game actions, and participant game actions (including explicitly provided feedback via keyboard presses). We also provide a CSV that provides demographic information for each participant. Additionally, the raw images of the participant during the games is available at github.com/yale-img/react.

### 3.3 Results

We first analyzed how the robot's visits affected human nonverbal signals as the data collection progressed. We used linear mixed models estimated with Restricted Maximum Likelihood (REML). The Game Number-Visit combination (e.g., Game3-First, Game4-Third, etc.) was a main effect and the participant ID was a random effect in the models. We conducted post-hoc Tukey Honestly Significant Difference (HSD) tests when appropriate.

We first examined the sum of AU activation values, as a proxy for participant expressiveness, during the robot visits in the interactions. Our analysis showed a significant difference by Game Number-Visit combination, $F(7, 7) = 16.54, p < 0.0001$. The post-hoc test revealed that the average of the sum of participant AU values during all three visits of both Game 3 and Game 4 were significantly higher than the robot's single visits in Games 5 and 6.

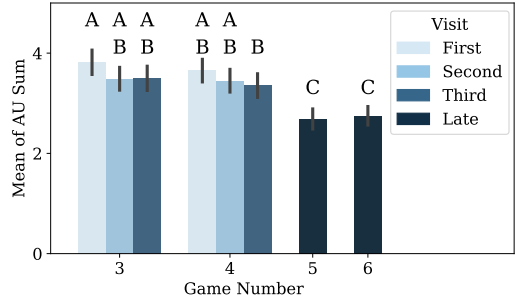

**Figure 2: Mean of sum of AU values during robot visits in REACT-Nao. Error bars are standard error. Letters (A,B,C) denote statistical significance. If visits do not share a letter, there is a statistically significant difference between values.**

Additionally, the average of the sum of participant AU values during the first visit of Game 3 was significantly higher than the third visit of Game 4. These differences between earlier and later visits show that participants reacted differently to similar stimuli based on when in the interaction they occurred. Figure 2 shows these results. A table of results is included in github.com/yale-img/react.

## 4 THE REACT-SHUTTER DATASET

REACT-Shutter contains data from interactions with a robot photographer [31]. A subset of this data was previously published [31], but it only included observations during specific robot actions. REACT-Shutter provides the complete interaction history, enabling better analyses and modeling.

### 4.1 Data Collection

First, participants consented to take part in the data collection, be video recorded, and have their data shared. Each participant then sat in front of a small robot while the robot took six photographs of them (as in Figure 1d). The robot, called Shutter, is a social robot with a screen face mounted on a small arm [2, 21]. Shutter took photos of the participants via a camera mounted on its head.

Each photograph was preceded by a series of four robot actions. These actions consisted of a mix of robot dialogue (telling jokes, telling the person to smile, and telling the person to relax) and changes to the robot's pose. The physical pose actions included aiming the robot's face directly at the participant, orienting its face away from the participant, or moving to one of four fixed poses. Actions were selected via weighted sampling, and an action could not be selected twice in a row – additional action details are included in the dataset documentation. Similar to Section 3.1, participants were not prompted to speak during the interactions.

Throughout the data collection, participants annotated robot actions based on their impressions of the robot's performance and answered survey questions. The whole interaction lasted between 45 minutes and one hour, and participants were compensated US$20. The protocol was approved by the local IRB. For more details about the data collection, please refer to Zhang et al. [31].

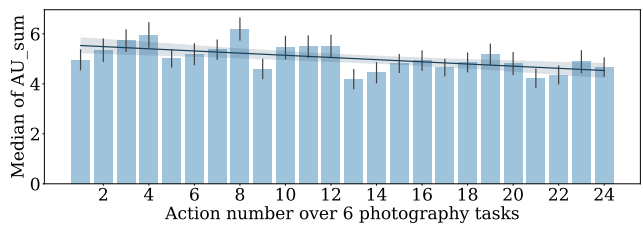

**Figure 3: Median of sum of AU values over the photography interaction. Error bars are standard error. Trend line is a linear regression model with a 95% confidence interval.**

### 4.2 Data Processing

The dataset consists of data collected for 40 participants, each of which completed six photography tasks.

**Facial Features.** The facial features were computed as in Section 3.2, but the data is organized into CSVs by photography task.

For our analyses, we first smoothed individual OpenFace features with a Gaussian window function, using the same approach as in Section 3.2. Additionally, we segmented the frames into action segments, splitting up the interaction based on when a new action began. We looked at the mean, median, maximum, and standard deviation of values of OpenFace features in each action segment. Post-processing scripts included in github.com/yale-img/react.

**Other features.** Our dataset includes additional information that provides context about the interaction. For each photography task, we include a CSV that provides the timestamps and details of robot actions throughout the task (e.g., specific utterance for a "joke" action). Additionally, we provide a summary CSV that provides additional information for each participant, including demographic information, the order of tasks, and the self-annotations. A full description of the features is available in the dataset documentation.

### 4.3 Results

We first explored how the expressiveness of participants changed over time as the interaction progressed. Considering all participants, we examined a variety of statistics (mean, median, max, standard deviation) over the sum of action unit activation values during the 24 actions that proceeded the individual photos in order. For example, see Figure 3 for the median values over each action segment.

For each statistic calculated over the sum of AU activation values during action segments, we employed a linear regression model to predict the statistic considering action number as the independent variable. Table A of the dataset documentation displays the results computed with the scipy.stats Python library [1]. Across all four summary statistics, there was a statistically significant negative slope, suggesting that participants became less expressive to robot actions over time. However, the slopes were just slightly negative, and the Pearson correlation coefficients were low suggesting that the model may not adequately capture the underlying relationships within the data. This is to be expected since expressivity likely depends on many other factors and warrants further study.

We fit another set of linear regression models, but this time considered whether the actions occurred first, second, third, or fourth in a mini-series before a photo as the independent variable. For these models, the slopes were positive for mean, median, and maximum values of the sum of action unit values over action segments (Table B of the dataset documentation). Taken with the previous results, this suggests that within a short photography task, participants got more expressive, but over time gradually became less expressive.

## 5 DISCUSSION

The REACT database has the potential to influence HRI work by facilitating research that examines automated reasoning about human reactions. This could enable a deeper understanding of the dynamics of human-robot interactions, which is essential for designing more effective robots. As we work towards enabling robots to help with physical and social tasks in human environments, it will be important to consider how novelty effects diminish and people change their responses to robots during interactions. Failing to account for changes in user expressivity could cause robots to fail to adjust their behavior to muted reactions later on in interactions.

As with all human subject data, there are ethical considerations [16] for the use of the REACT database. Responsible use guidelines include ensuring that the data is not used for purposes that would negatively manipulate or impact people.

Our database facilitates exciting research directions but it is not without limitations. The datasets showcase interactions for two different tasks, allowing users to explore model generalizability; however, it is unclear how analyses or models specific to these two tasks would translate to other interaction scenarios. Also, there are other forms of implicit communicative signals, such as the tone of verbal communications, that are not included in the datasets.

## 6 CONCLUSION

We contributed two datasets that can facilitate studying how robots can improve their behavior based on naturalistic human reactions. Additionally, we found preliminary evidence highlighting the importance of considering the interaction history when interpreting human reactions in HRI. We hope that the REACT database and initial findings encourage the HRI community to further explore how robots can learn from implicit human feedback over time.

## ACKNOWLEDGMENTS

This work was partly funded by the National Science Foundation (under Grant No. IIS-1924802, IIS-2106690, and IIS-1955653), Tata Sons Private Limited, Tata Consultancy Services Limited, and Titan. H. Zhou was funded by the Andy Keidel Fund and the Yale College Dean's Research Fellowship. S. Richardson was funded by the Yale College First-Year Summer Research Fellowship.

## REFERENCES

[1] [n. d.]. SciPy linregress Function Documentation. https://docs.scipy.org/doc/scipy/reference/generated/scipy.stats.linregress.html. Accessed October 1, 2023.
[2] Timothy Adamson, C Burton Lyng-Olsen, Kendrick Umstattd, and Marynel Vázquez. 2020. Designing social interactions with a humorous robot photographer. In *Proceedings of the 2020 ACM/IEEE International Conference on Human-Robot Interaction*. 233–241.
[3] Gopika Ajaykumar, Maureen Steele, and Chien-Ming Huang. 2021. A survey on end-user robot programming. *ACM Computing Surveys (CSUR)* 54, 8 (2021), 1–36.

REACT: Two Datasets for Analyzing Both Human Reactions and Evaluative Feedback to Robots Over Time[4] Tadas Baltrusaitis, Amir Zadeh, Yao Chong Lim, and Louis-Philippe Morency. 2018. Openface 2.0: Facial behavior analysis toolkit. In *2018 13th IEEE international conference on automatic face & gesture recognition (FG 2018)*. IEEE, 59–66.

[5] Lisa Feldman Barrett, Ralph Adolphs, Stacy Marsella, Aleix M Martinez, and Seth D Pollak. 2019. Emotional expressions reconsidered: Challenges to inferring emotion from human facial movements. *Psychological science in the public interest* 20, 1 (2019), 1–68.

[6] Atef Ben-Youssef, Chloé Clavel, Slim Essid, Miriam Bilac, Marine Chamoux, and Angelica Lim. 2017. UE-HRI: A New Dataset for the Study of User Engagement in Spontaneous Human-Robot Interactions. In *Proceedings of the 19th ACM International Conference on Multimodal Interaction* (Glasgow, UK) *(ICMI '17)*. Association for Computing Machinery, New York, NY, USA, 464–472. https://doi.org/10.1145/3136755.3136814

[7] Erdem Bıyık, Aditi Talati, and Dorsa Sadigh. 2022. Aprel: A library for active preference-based reward learning algorithms. In *2022 17th ACM/IEEE International Conference on Human-Robot Interaction (HRI)*. IEEE, 613–617.

[8] Alexandra Bremers, Maria Teresa Parreira, Xuanyu Fang, Natalie Friedman, Adolfo Ramirez-Aristizabal, Alexandria Pabst, Mirjana Spasojevic, Michael Kuniavsky, and Wendy Ju. 2023. The Bystander Affect Detection (BAD) Dataset for Failure Detection in HRI. *arXiv preprint arXiv:2303.04835* (2023).

[9] Kate Candon, Jesse Chen, Yoony Kim, Zoe Hsu, Nathan Tsoi, and Marynel Vázquez. 2023. Nonverbal Human Signals Can Help Autonomous Agents Infer Human Preferences for Their Behavior. In *Proceedings of the 2023 International Conference on Autonomous Agents and Multiagent Systems*. 307–316.

[10] Kate Candon, Helen Zhou, Sarah Gillet, and Marynel Vázquez. 2023. Verbally Soliciting Human Feedback in Continuous Human-Robot Collaboration: Effects of the Framing and Timing of Reminders. In *Proceedings of the 2023 ACM/IEEE International Conference on Human-Robot Interaction*. 290–300.

[11] Oya Celiktutan, Efstratios Skordos, and Hatice Gunes. 2017. Multimodal human-human-robot interactions (mhhri) dataset for studying personality and engagement. *IEEE Transactions on Affective Computing* 10, 4 (2017), 484–497.

[12] Nikhil Churamani, Sinan Kalkan, and Hatice Gunes. 2020. Continual Learning for Affective Robotics: Why, What and How?. In *2020 29th IEEE International Conference on Robot and Human Interactive Communication (RO-MAN)*. IEEE, 425–431.

[13] Yuchen Cui, Qiping Zhang, Alessandro Allievi, Peter Stone, Scott Niekum, and W. Bradley Knox. 2020. The EMPATHIC Framework for Task Learning from Implicit Human Feedback. In *CoRL*.

[14] Yuchen Cui, Qiping Zhang, Brad Knox, Alessandro Allievi, Peter Stone, and Scott Niekum. 2021. The empathic framework for task learning from implicit human feedback. In *Conference on Robot Learning*. PMLR, 604–626.

[15] Terrence Fong, Illah Nourbakhsh, and Kerstin Dautenhahn. 2003. A survey of socially interactive robots. *Robotics and autonomous systems* 42, 3-4 (2003), 143–166.

[16] Isabelle Hupont, Songül Tolan, Hatice Gunes, and Emilia Gómez. 2022. The landscape of facial processing applications in the context of the European AI Act and the development of trustworthy systems. *Scientific Reports* 12, 1 (2022), 10688.

[17] Rachael E Jack, Oliver GB Garrod, Hui Yu, Roberto Caldara, and Philippe G Schyns. 2012. Facial expressions of emotion are not culturally universal. *Proceedings of the National Academy of Sciences* 109, 19 (2012), 7241–7244.

[18] Dinesh Babu Jayagopi, Samira Sheiki, David Klotz, Johannes Wienke, Jean-Marc Odobez, Sebastien Wrede, Vasil Khalidov, Laurent Nyugen, Britta Wrede, and Daniel Gatica-Perez. 2013. The vernissage corpus: A conversational Human-Robot-Interaction dataset. In *2013 8th ACM/IEEE International Conference on Human-Robot Interaction (HRI)*. 149–150. https://doi.org/10.1109/HRI.2013.6483545

[19] Adam Kendon. 1990. *Conducting interaction: Patterns of behavior in focused encounters*. Vol. 7. CUP Archive.

[20] Iolanda Leite, André Pereira, Carlos Martinho, Ana Paiva, and Ginevra Castellano. 2009. Towards an empathic chess companion. In *Proceedings of the 8th International Conference on Autonomous agents and Multiagent Systems (AAMAS 2009)*. 33–36.

[21] Alexander Lew, Sydney Thompson, Nathan Tsoi, and Marynel Vázquez. 2023. Shutter, the Robot Photographer: Leveraging Behavior Trees for Public, In-the-Wild Human-Robot Interactions. *arXiv preprint arXiv:2302.00191* (2023).

[22] Guangliang Li, Hamdi Dibeklioğlu, Shimon Whiteson, and Hayley Hung. 2020. Facial feedback for reinforcement learning: a case study and offline analysis using the TAMER framework. *Autonomous Agents and Multi-Agent Systems* 34, 1 (2020), 1–29.

[23] James MacGlashan, Mark K. Ho, Robert Loftin, Bei Peng, Guan Wang, David L. Roberts, Matthew E. Taylor, and Michael L. Littman. 2017. Interactive Learning from Policy-Dependent Human Feedback. In *Proceedings of the 34th International Conference on Machine Learning (Proceedings of Machine Learning Research, Vol. 70)*, Doina Precup and Yee Whye Teh (Eds.). PMLR, 2285–2294. https://proceedings.mlr.press/v70/macglashan17a.html

[24] Benjamin A Newman, Reuben M Aronson, Siddhartha S Srinivasa, Kris Kitani, and Henny Admoni. 2022. HARMONIC: A multimodal dataset of assistive human–robot collaboration. *The International Journal of Robotics Research* 41, 1 (2022), 3–11.

[25] Benjamin A Newman, Christopher Jason Paxton, Kris Kitani, and Henny Admoni. 2023. Towards Online Adaptation for Autonomous Household Assistants. In *Companion of the 2023 ACM/IEEE International Conference on Human-Robot Interaction*. 506–510.

[26] Rosalind W Picard. 2000. *Affective computing*. MIT press.

[27] Niyati Rawal and Ruth Maria Stock-Homburg. 2022. Facial emotion expressions in human–robot interaction: A survey. *International Journal of Social Robotics* 14, 7 (2022), 1583–1604.

[28] Maia Stiber, Russell H. Taylor, and Chien-Ming Huang. 2023. On Using Social Signals to Enable Flexible Error-Aware HRI. In *Proceedings of the 2023 ACM/IEEE International Conference on Human-Robot Interaction* (Stockholm, Sweden) *(HRI '23)*. Association for Computing Machinery, New York, NY, USA, 222–230. https://doi.org/10.1145/3568162.3576990

[29] Alessandro Vinciarelli, Maja Pantic, and Hervé Bourlard. 2009. Social signal processing: Survey of an emerging domain. *Image and vision computing* 27, 12 (2009), 1743–1759.

[30] Morton Wiener, Shannon Devoe, Stuart Rubinow, and Jesse Geller. 1972. Nonverbal behavior and nonverbal communication. *Psychological review* 79, 3 (1972), 185.

[31] Qiping Zhang, Austin Narcomey, Kate Candon, and Marynel Vázquez. 2023. Self-Annotation Methods for Aligning Implicit and Explicit Human Feedback in Human-Robot Interaction. In *Proceedings of the 2023 ACM/IEEE International Conference on Human-Robot Interaction* (Stockholm, Sweden) *(HRI '23)*. Association for Computing Machinery, New York, NY, USA, 398–407. https://doi.org/10.1145/3568162.3576986
*Preprint*